\PassOptionsToPackage{table}{xcolor} 

\documentclass[format=sigconf,anonymous=false,review=false]{acmart}

\usepackage{amsmath,amssymb,amsfonts}
\usepackage[ruled,vlined]{algorithm2e}
\usepackage{graphicx}
\usepackage{textcomp}
\usepackage{xcolor}
\usepackage{libertine}
\renewcommand{\abstractname}{\uppercase{Abstract}}

\AtBeginDocument{%
  }

\acmDOI{nn.nnnn/nnnnnnn.nnnnnnn} 
\acmISBN{xxx-x-xxxx-xxxx-x/YY/MM} 
\acmConference[GECCO '25]{The Genetic and Evolutionary Computation Conference 2025}{July 14--18, 2025}{Málaga, Spain}
\acmYear{2025}
\copyrightyear{2025}

\begin{document}

\title{Residual Learning Inspired Crossover Operator and Strategy Enhancements for Evolutionary Multitasking}

\author{Ruilin Wang}
\email{wrl61@mail.ecust.edu.cn}
\affiliation{%
  \institution{Department of Computer Science and Engineering, East China University of Science and Technology}
  \city{Shanghai}
  \country{China}
}

\author{Xiang Feng}
\authornote{*Corresponding author.}
\email{xfeng@ecust.edu.cn}
\affiliation{%
  \institution{Department of Computer Science and Engineering, East China University of Science and Technology}
  \city{Shanghai}
  \country{China}
}

\author{Huiqun Yu}
\email{yhq@ecust.edu.cn}
\affiliation{%
  \institution{Department of Computer Science and Engineering, East China University of Science and Technology}
  \city{Shanghai}
  \country{China}
}

\author{Edmund M-K Lai}
\email{edmund.lai@aut.ac.nz}
\affiliation{%
  \institution{Department of Data Science and Artificial Intelligence, Auckland University of Technology}
  \city{Auckland}
  \country{New Zealand}
}

\renewcommand{\shortauthors}{Ruilin et al.}

\begin{abstract}
In evolutionary multitasking, strategies such as crossover operators and skill factor assignment are critical for effective knowledge transfer. Existing improvements to crossover operators primarily focus on low-dimensional variable combinations, such as arithmetic crossover or partially mapped crossover, which are insufficient for modeling complex high-dimensional interactions.Moreover, static or semi-dynamic crossover strategies fail to adapt to the dynamic dependencies among tasks. In addition, current Multifactorial Evolutionary Algorithm frameworks often rely on fixed skill factor assignment strategies, lacking flexibility. To address these limitations, this paper proposes the Multifactorial Evolutionary Algorithm-Residual Learning (MFEA-RL) method based on residual learning. The method employs a Very Deep Super-Resolution (VDSR) model to generate high-dimensional residual representations of individuals, enhancing the modeling of complex relationships within dimensions. A ResNet-based mechanism dynamically assigns skill factors to improve task adaptability, while a random mapping mechanism efficiently performs crossover operations and mitigates the risk of negative transfer. Theoretical analysis and experimental results show that MFEA-RL outperforms state-of-the-art multitasking algorithms. It excels in both convergence and adaptability on standard evolutionary multitasking benchmarks, including CEC2017-MTSO and WCCI2020-MTSO. Additionally, its effectiveness is validated through a real-world application scenario.   
\end{abstract}

\ccsdesc[500]{Computing methodologies~Artificial intelligence}
\ccsdesc[500]{Theory of computation~Evolutionary algorithms}
\keywords{Multifactorial evolutionary algorithm, knowledge transfer, crossover operator, residual learning, theoretical analysis }
\maketitle

\section{INTRODUCTION}
With the increasing prevalence of multitasking optimization problems in practical applications, efficiently sharing knowledge across tasks to enhance overall optimization performance has become a critical research focus. Evolutionary Multitasking (EMT) \cite{ong_emt} integrates multiple optimization tasks into a unified algorithmic framework by sharing populations and genetic operations. This approach not only significantly reduces computational costs but also leverages inter-task correlations to improve search efficiency \cite{lda_mfea,liaw}. In EMT, crossover operators and skill factor assignment strategies are central to enabling effective knowledge transfer and task collaboration. However, existing methods face significant challenges in handling complex inter-task relationships \cite{mingf_1, mingf_2, li_gecco}.

\begin{figure*}[h!]
\centerline{\includegraphics[width=7.2in]{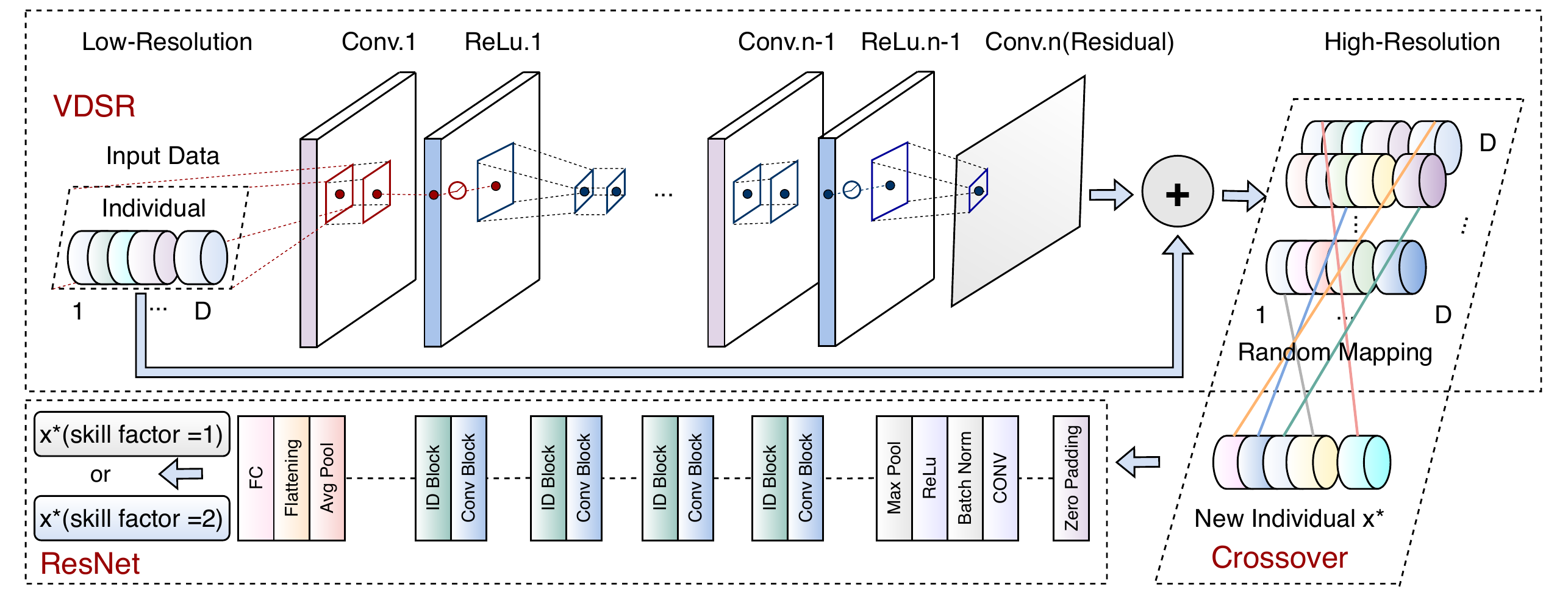}}
\caption{Architecture of MFEA-RL with residual learning and random mapping}
\label{architecture}
\end{figure*}

While existing crossover optimization methods exhibit competitive performance, they primarily focus on low-dimensional variable combinations (e.g., arithmetic or partially mapped crossover), making them inadequate for modeling complex interactions in high-dimensional tasks, especially under nonlinear or highly correlated conditions \cite{wang_gecco, zhang_gecco}. Moreover, static or semi-dynamic crossover operators, though effective for certain tasks, struggle to adapt to the dynamic dependencies among tasks. Additionally, current MFEA frameworks often rely on fixed skill factor assignment strategies, which lack flexibility. 

To address these issues, we propose the MFEA-RL method based on the concept of residual learning. First, a 1×$D$ individual, where $D$ represents the dimensions of decision variables, is transformed into high-dimensional residual data via a Very Deep Super-Resolution (VDSR) model \cite{vdsr}, which is combined with the input individual to generate a $D$×$D$ high-dimensional representation, enabling better modeling of complex relationships. Next, ResNet \cite{resnet} is utilized to dynamically assign skill factors, integrating high-dimensional residual information and task relationship learning to optimize individual adaptability across tasks. Finally, random mapping is employed to extract a single row from the high-dimensional data and map it back to the 1×$D$ space, completing the crossover operation and replacing traditional simulated binary crossover (SBX). The proposed MFEA-RL method explicitly models variable interactions through high-dimensional representations, overcoming the limitations of existing methods in high-dimensional correlation modeling. Residual learning enhances feature extraction capabilities, enabling the model to better handle complex tasks. Random mapping introduces dynamic adjustment capabilities, reducing the risk of negative transfer and improving the robustness of knowledge transfer across tasks. Our contributions can be summarized as follows:

\begin{itemize}
\item[1)]
We propose a novel crossover operator that leverages high-dimensional residual representations generated by a VDSR model. By transforming low-dimensional individuals into high-dimensional spaces and integrating residual data with the original input, the operator explicitly models complex inter-variable relationships. This approach enhances the crossover's ability to preserve and exploit critical task-specific information, enabling more effective optimization in tasks with high-dimensional correlations.
\end{itemize}

\begin{itemize}
\item[2)]
We design a ResNet-based mechanism for dynamic skill factor assignment, which leverages high-dimensional residual information and inter-task relationship learning. This approach allows us to adaptively assign skill factors, addressing the limitations of static or semi-dynamic strategies in traditional multifactorial evolutionary algorithms.
\end{itemize}

\begin{itemize}
\item[3)]
Through theoretical analysis and extensive experimental evaluations, we demonstrate the effectiveness of our proposed framework. The MFEA-RL method achieves outstanding convergence speed and solution quality on standard multitasking optimization benchmarks, outperforming several state-of-the-art approaches.
\end{itemize}

\section{PRELIMINARYR}
\subsection{Multifactorial Evolutionary Algorithm}

In evolutionary multitasking, the goal is to solve multiple optimization problems simultaneously. Suppose we are given \( T \) tasks, each formulated as a single-objective minimization problem. For task \( T_i \), its decision space is denoted by \( X_i \) and the corresponding objective function is \( f_i: X_i \rightarrow \mathbb{R} \). The multitasking problem aims to find the optimal solutions \( x^*_i \in X_i \) that minimize each task’s objective independently, expressed as:
\begin{align}
\{x^*_1, x^*_2, \dots, x^*_T\} = \Big\{ 
&\arg\min_{x_1} f_1(x_1), 
\arg\min_{x_2} f_2(x_2), \nonumber \\
&\dots, 
\arg\min_{x_T} f_T(x_T) 
\Big\}
\end{align}

These tasks may differ in dimensionality, search space characteristics, and fitness landscapes, making joint optimization particularly challenging yet beneficial for knowledge transfer across tasks.

The MFEA offers a unified framework for solving such multitasking problems by optimizing a shared population across all tasks. It leverages genetic operators and implicit knowledge transfer to facilitate cross-task learning. A central concept in MFEA is the skill factor, which denotes the task that an individual is currently specialized in. Each individual is associated with one skill factor, and during the evolutionary process, it only contributes its fitness evaluation to the corresponding task. This task-specific assignment allows the algorithm to implicitly preserve task-related knowledge while enabling occasional crossovers between individuals with different skill factors. While MFEA also involves auxiliary definitions like factorial cost, factorial rank, and scalar fitness to evaluate individuals across tasks, the skill factor serves as the primary mechanism guiding task assignment and knowledge transfer, making it crucial to the algorithm’s multitasking capability.

\subsection{Related Work}
In EMT, effective knowledge transfer and task collaboration play a critical role in improving optimization performance. Existing studies have explored multiple directions to enhance knowledge transfer and operator design. Molaei et al.~\cite{psolc} introduced a learning approach combining particle swarm optimization with arithmetic crossover, effectively enhancing global search capability while mitigating premature convergence. Koohestani~\cite{pmx} proposed the Partially Mapped Crossover (PMX), which is tailored for permutation-based problems and improves solution quality. Li et al.~\cite{sao} developed the MTEA-SaO algorithm, incorporating solver adaptation and implicit knowledge transfer to dynamically select optimizers across tasks. Zhou et al.~\cite{mfea_akt} designed a dynamic crossover strategy that adapts to evolving knowledge transfer requirements, addressing the limitations of static crossover approaches. Liu et al.~\cite{olc} derived a linear crossover operator to optimize parameters, reducing the risk of negative transfer effects.

Xue et al.~\cite{at_mfea} introduced an affine transformation-based domain adaptation technique, significantly improving inter-task mapping for heterogeneous multitasking scenarios. Additionally, subspace alignment methods~\cite{ascmfde} and block-level knowledge transfer frameworks~\cite{blktde} have been developed to align task-specific subspaces and exploit relationships across tasks, enhancing optimization efficiency. Bull et al.~\cite{bull} proposed a global crossover method inspired by cooperative coevolution, enhancing inter-task collaboration through global information sharing. Furthermore, Bai et al.~\cite{mtes} extended multitasking optimization using a gradient descent framework, demonstrating that multitasking can achieve faster convergence compared to single-task optimization under specific conditions.

Recently, neural networks have been increasingly adopted in EMT to model variable dependencies and enable efficient solution mapping across tasks. Wang et al.~\cite{wrl_transfer} and related studies~\cite{ann} demonstrated that neural architectures can capture task-specific relationships and improve solution quality, particularly in scenarios with high-dimensional variable interactions. These findings highlight the potential of neural networks in addressing multitasking optimization problems.

\section{PROPOSED METHOD}

\subsection{MFEA-RL}
Existing EMT algorithms improve performance by transferring knowledge across tasks but face challenges in managing dynamic task relationships, mitigating negative transfer, and modeling complex variable interactions. The proposed MFEA-RL algorithm addresses these limitations by integrating residual learning and dynamic skill factor assignment. As shown in Figure \ref{architecture}, a VDSR model transforms low-dimensional individuals $I_l \in \mathbb{R}^{1 \times D}$ into high- dimensional representations $I_h \in \mathbb{R}^{D \times D}$, enabling explicit modeling of complex variable interactions. This transformation preserves the original input information through residual connections while capturing additional features for multi-task optimization. The algorithm employs a pre-trained ResNet, trained on a large dataset of individuals with skill factor attributes, to dynamically assign skill factors. By leveraging learned task-specific dependencies, the ResNet effectively classifies individuals into different task categories, allowing adaptive allocation of computational resources. To mitigate negative transfer, the algorithm incorporates a random mapping mechanism during the crossover operation. A single row from the high-dimensional residual representation is mapped back to the original \(1 \times D\) space, replacing the traditional SBX operator. This mechanism utilizes high-dimensional features to selectively transfer information, reducing unintended interference and promoting efficient task collaboration. Additionally, it improves diversity maintenance, ensuring robust optimization across multiple tasks.

\begin{algorithm}
\caption{MFEA-RL Framework}
\label{alg:mfea-rl}
\KwIn{Problem set $\mathcal{F} = \{\mathcal{F}_1, \dots, \mathcal{F}_T\}$ with $T$ tasks, population size $N$, decision variable dimension $D$, VDSR network $\mathcal{N}_\text{VDSR}$, ResNet network $\mathcal{N}_\text{ResNet}$, random mating probability $RMP$, mutation rate $\mu_M$}
\KwOut{Optimized solutions $\mathcal{S} = \{\mathbf{x}_1^*, \dots, \mathbf{x}_T^*\}$}
\BlankLine

\textbf{Initialize:} Population $\mathcal{P}_0 = \{\mathbf{x}_i\}_{i=1}^N \in \mathbb{R}^{N \times D}$\;
\While{$epoch$ < $MaxEpoch$}{
    Extract $\mathcal{D}_\text{Dec}$, $\mathcal{F}_\text{Obj}$, and  $\mathcal{F}_\text{Task}$ from $\mathcal{P}_t$\;
    $I_h=\mathcal{N}_\text{VDSR}(I_l)$ \;

    \ForEach{$(\mathbf{x}_1, \mathbf{x}_2) \in \mathcal{P}_t$}{
        \eIf{$\mathcal{F}_\text{Task}(\mathbf{x}_1) = \mathcal{F}_\text{Task}(\mathbf{x}_2)$ \textbf{or} $\text{rand}() < \text{RMP}$}{
            Randomly select row $\mathbf{r} \in I_h$ and map $\mathbf{r} \to \mathbb{R}^{1 \times D}$\;
            Assign task labels: $\mathcal{F}_\text{Task}=\mathcal{N}_\text{ResNet}(I_h)$\;
        }{
            Mutation: $\mathbf{x}'_1=\mathbf{x}_1(\mu_M)$, $\mathbf{x}'_2=\mathbf{x}_2(\mu_M)$\;
        }
        Correct boundary violations for offspring $\mathbf{x}^\prime$\;
    }
    Evaluate $\mathbf{x}^\prime$ on all tasks in $\mathcal{P}$ to update $\mathcal{F}_\text{Obj}$\;
    Select top-performing individuals to form $\mathcal{P}_{t+1}$\ based on scalar fitness ranking;
}
\textbf{Output:} Optimized solutions $\mathcal{S}$\;
\end{algorithm}

\begin{algorithm}
\caption{Training Process of VDSR and ResNet Models in the Algorithm Architecture}
\label{alg:training-framework}
\KwIn{
Low-resolution training data $\mathbf{X}$, 
number of epochs $\text{numEpochs}_1$ (VDSR) and $\text{numEpochs}_2$ (ResNet), 
training options $\mathcal{T}$, including learning rate, optimizer, and regularization terms, 
mini-batch size $B$, 
VDSR network $\mathcal{N}_\text{VDSR}$, 
ResNet architecture $\mathcal{N}_\text{ResNet}$ with modified input and output layers
}
\KwOut{
Integrated network $\mathcal{N}$ (VDSR with ResNet)
}
\BlankLine

\textbf{Part 1: Train VDSR Model} \\
\Indp
\For{$t = 1$ \textbf{to} $\text{numEpochs}_1$}{
    Shuffle $\mathbf{X}$\;
    \ForEach{mini-batch $\mathbf{X}_\text{batch} \subset \mathbf{X}$ of size $B$}{
        Residuals: $\mathbf{R} \leftarrow \mathcal{N}_\text{VDSR}(\mathbf{X}_\text{batch})$\;
        Predictions: $\hat{\mathbf{X}} \leftarrow \mathbf{X}_\text{batch} + \mathbf{R}$\;
        Loss: $\mathcal{L}_1 \leftarrow \frac{1}{B} \sum (\hat{\mathbf{X}} - \mathbf{X}_\text{batch})^2$\;
        Update $\mathcal{N}_\text{VDSR}$ using $\mathcal{L}_1$\;
    }
}
\textbf{Output:} $\mathcal{N}_\text{VDSR'}$\;
\Indm

\BlankLine
\textbf{Part 2: Train ResNet Model} \\
\Indp
Split $\mathbf{X}$ into training ($\mathcal{X}_\text{train}$) and validation ($\mathcal{X}_\text{val}$) sets\;
\For{$t = 1$ \textbf{to} $\text{numEpochs}_2$}{
    Shuffle $\mathcal{X}_\text{train}$\;
    \ForEach{mini-batch $\mathbf{X}_\text{batch} \subset \mathcal{X}_\text{train}$ of size $B$}{
        Outputs: $\hat{\mathbf{Y}} \leftarrow \mathcal{N}_\text{ResNet}(\mathbf{X}_\text{batch})$\;
        Loss: $\mathcal{L}_2 \leftarrow \text{CrossEntropy}(\hat{\mathbf{Y}}, \mathbf{Y}_\text{batch})$\;
        Update $\mathcal{N}_\text{ResNet}$ using $\mathcal{L}_2$\;
    }
    Evaluate validation accuracy $\text{Accuracy}_\text{val}(t)$ on $\mathcal{X}_\text{val}$\;
    \If{$\text{Accuracy}_\text{val}(t) \leq \max(\text{Accuracy}_\text{val}(t-p:t-1))$ 
    for $p$ consecutive epochs}{
    Terminate training\;
}
}
\textbf{Output:} $\mathcal{N}_\text{ResNet}$\;
\Indm

\end{algorithm}

Algorithm \ref{alg:training-framework} outlines the training process for the VDSR and ResNet models and their integration into the MFEA-RL framework. In MFEA-RL, these networks are trained using data generated from the evolving population without requiring additional function evaluations. At each generation, evaluated individuals are sampled and grouped by task labels to construct balanced training sets for skill factor prediction. Each $1 \times D$ individual serves as a low-resolution input, while the corresponding high-dimensional target is constructed using augmented data within the same task group. The VDSR network $\mathcal{N}_\text{VDSR}$ is trained to predict high-dimensional residuals $\mathbf{R} \in \mathbb{R}^{D \times D}$, which are added to the original input via element-wise addition to form $\hat{\mathbf{X}} = \mathbf{X}_\text{batch} + \mathbf{R}$, where $\mathbf{X}_\text{batch} \in \mathbb{R}^{1 \times D}$ is broadcasted accordingly. This residual formulation enhances variable dependencies without implying the model is trained to minimize the residual to zero. The network parameters are updated over multiple epochs using mean squared error loss $\mathcal{L}_1$, and the trained VDSR model $\mathcal{N}_\text{VDSR'}$ is retained. Subsequently, the ResNet model $\mathcal{N}_\text{ResNet}$ is trained to predict skill factor labels using the residual-enhanced data. The data are split into training and validation sets, optimized via cross-entropy loss $\mathcal{L}_2$, and early stopping is applied based on validation accuracy. The resulting model $\mathcal{N}_\text{ResNet'}$ is used during evolution to dynamically assign task labels to offspring based on learned task-specific representations. Throughout this paper, we use $D$ to denote the number of decision variables and $T$ to denote the number of tasks.

With the VDSR and ResNet networks trained, the MFEA-RL optimization process proceeds as shown in Algorithm \ref{alg:mfea-rl}. The algorithm begins by initializing a population of size $N$, with individuals represented as $\mathbf{x} \in \mathbb{R}^{1 \times D}$. Each individual's decision variables $D_{\text{Dec}}$, objective values $F_{\text{Obj}}$, and task labels $F_{\text{Task}}$ are extracted for further processing. The trained VDSR model maps these $1 \times D$ vectors into $D \times D$ high-dimensional residual representations to enable complex inter-variable modeling.For each pair of parents selected from the population, if they belong to the same task or satisfy the random mating probability, the algorithm performs a random mapping by selecting a single row $\mathbf{r} \in \mathbb{R}^{1 \times D}$ from the high-dimensional residual representation. This row serves as a task-independent latent representation and is used to replace one parent in the crossover operation, enabling implicit knowledge transfer across tasks. This replaces the traditional SBX operator and helps mitigate negative transfer by leveraging high-dimensional context. The ResNet then assigns updated task labels to the offspring based on their transformed features. If the parents belong to different tasks and the random mating condition is not satisfied, polynomial mutation is applied to maintain diversity. Boundary violations are corrected to ensure feasibility, and offspring are evaluated across tasks. Their fitness is used to select the top-performing individuals for the next generation. This process repeats until a termination criterion is reached, such as a maximum number of generations or performance convergence. The algorithm ultimately returns optimized solutions for all tasks.

\subsection{Theoretical Analysis}
In this section, we analyze the proposed method by incorporating the insights of the VDSR model, leveraging high-dimensional representations, random mapping, and variable interaction modeling. The analysis demonstrates how the method reduces the likelihood of negative transfer while enhancing the diversity of the search space. Specifically, we provide a rigorous proof using the Johnson-Lindenstrauss (JL) Lemma \cite{jl} to show that random mapping approximately preserve pairwise distances, ensuring the integrity of variable relationships.

Consider $K$ optimization tasks, where each task $T_j$ is defined as a single-objective minimization problem:
\begin{equation}
    \hat{x}_j = \arg\min_{x_j \in X_j} f_j(x_j), \quad j = 1, 2, \dots, K,
\end{equation}
where $X_j \subseteq \mathbb{R}^D$ is the search space for task $T_j$, $f_j: X_j \to \mathbb{R}$ is the objective function, and $\hat{x}_j$ is the optimal solution for task $T_j$. 

Our proposed method employs a VDSR model to transform low-dimensional individuals into high-dimensional representations during the crossover process. The high-dimensional data is then randomly mapped back to its original size before being crossed with the original individual. Let $x \in \mathbb{R}^{1 \times D}$ represent a low-dimensional individual. The VDSR model maps $x$ to a high-dimensional matrix:
\begin{equation}
    X_{\text{new}} = x + R, \quad R \in \mathbb{R}^{D \times D}
\end{equation}
Here, the operator “$+$” denotes row-wise broadcasting, where the $1 \times D$ vector $x$ is replicated $D$ times to form a matrix of shape $D \times D$ before being added to the residual matrix $R$. This operation ensures dimension alignment and allows the model to incorporate both the original variable information and the learned high-order residuals during the crossover process.

The expanded search space is defined as:
\begin{equation}
    X^{\text{high}} = \{X_{\text{new}} \mid X_{\text{new}} = x + R, \, R = \mathcal{F}(x; \Theta), \, x \in X\},
\end{equation}
where $\mathcal{F}(\cdot; \Theta)$ is the residual learning function parameterized by $\Theta$. The inclusion of $R$ allows the model to explore a significantly larger search space:
\begin{equation}
    \frac{|X^{\text{high}}|}{|X|} \propto d^{D},
\end{equation}
where $|X|$ and $|X^{\text{high}}|$ are the volumes of the original and expanded search spaces, respectively. This expanded search space provides increased diversity, which is crucial for avoiding premature convergence.

After generating the high-dimensional matrix $X_{\text{new}}$, the method applies random mapping to extract meaningful subsets for evaluation. A random mapping is performed by selecting a random row from $X_{\text{new}}$, resulting in:
\begin{equation}
    x_{\text{proj}} = P X_{\text{new}},
\end{equation}
where $P \in \mathbb{R}^{1 \times D}$ is a random mapping matrix. This process can be interpreted as reducing the high-dimensional representation to a manageable subset while retaining critical information.

The JL Lemma guarantees that random mapping approximately preserve pairwise distances between points in high- dimensional spaces. For a set of $n$ points $\{X_{\text{new}}^{(1)}, X_{\text{new}}^{(2)}, \dots, X_{\text{new}}^{(n)}\} \subset \mathbb{R}^{D \times D}$, there exists a random mapping $f: \mathbb{R}^D \to \mathbb{R}^k$ such that for any two points $X_{\text{new}}^{(i)}$ and $X_{\text{new}}^{(j)}$:
\begin{equation}
\begin{aligned}
    &(1 - \epsilon) \|X_{\text{new}}^{(i)} - X_{\text{new}}^{(j)}\|^2\\
    &\leq \|f(X_{\text{new}}^{(i)}) - f(X_{\text{new}}^{(j)})\|^2 \\
    &\leq (1 + \epsilon) \|X_{\text{new}}^{(i)} - X_{\text{new}}^{(j)}\|^2,
\end{aligned}
\end{equation}
where $\epsilon \in (0, 1)$ is the approximation error, and $k = O(\ln n / \epsilon^2)$ is the reduced dimensionality.

Let $X_{\text{new}}^{(i)}, X_{\text{new}}^{(j)} \in \mathbb{R}^{D \times D}$ be two high-dimensional points, and let $v = X_{\text{new}}^{(i)} - X_{\text{new}}^{(j)}$. The Frobenius norm of their difference in the original space is:
\begin{equation}
    \|X_{\text{new}}^{(i)} - X_{\text{new}}^{(j)}\|_F^2 = \sum_{k=1}^{D} \sum_{l=1}^{D} v_{kl}^2.
\end{equation}

After applying a random mapping $P$, the projected distance is:
\begin{equation}
    \|f(X_{\text{new}}^{(i)}) - f(X_{\text{new}}^{(j)})\|^2 = \|P v\|^2.
\end{equation}

From the properties of random mapping matrices, we have:
\begin{equation}
    \mathbb{E}[\|P v\|^2] = \|v\|^2, \quad \text{Var}(\|P v\|^2) \leq \epsilon \|v\|^2,
\end{equation}
where \(\mathbb{E}[\cdot]\) denotes the expected value, representing the mean of the random variable over all possible outcomes. Thus:

\begin{equation}
    (1 - \epsilon) \|v\|^2 \leq \|P v\|^2 \leq (1 + \epsilon) \|v\|^2.
\end{equation}
This ensures approximate preservation of pairwise distances, reducing the likelihood of misleading projections.

The VDSR model encodes variable relationships in \(X_{\text{new}}\), and random mapping ensure that the encoded relationships are preserved during dimensionality reduction. By maintaining pairwise distances, the projected point \(x_{\text{proj}}\) remains close to task-specific optima, mitigating the risk of negative transfer:
\begin{equation}
    \mathbb{P}\left[\|x_{\text{proj}} - \hat{x}_j\| \leq \|X_{\text{new}} - \hat{X}_j\|\right] \geq 1 - \delta,
\end{equation}
where \(\mathbb{P}[\cdot]\) represents the probability of the event occurring, and \(\delta\) is the confidence level.

The integration of VDSR for residual learning, combined with random mapping and the JL Lemma, ensures the effective modeling of variable interactions while preserving relationships across tasks. This reduces the risk of negative transfer and enhances multitask optimization performance.

\section{EXPERIMENTAL STUDIES}
\subsection{Experimental Configuration}
All experiments follow standard settings commonly used in the EMT domain \cite{wrl_transfer}. The population size is set to 100 for both the proposed method and all baselines. For SBX-based algorithms, the crossover distribution index is 2, the mutation index is 5, and the random mating probability is 0.3. The proposed method employs a VDSR model with 64 hidden channels, a learning rate of 0.001, batch size of 20, and 5 training epochs. A custom training loop using the Adam optimizer is adopted, with $\beta_1 = 0.9$ and $\beta_2 = 0.999$. The ResNet model is trained with a learning rate of 0.001, batch size of 32, and 50 epochs, using $\beta_1 = 0.9$, $\beta_2 = 0.9999$, one input channel, and a maximum time step of 1,000.

\begin{table*}[h!]
\vspace{1em}
    \setlength{\tabcolsep}{1.5pt}
    \renewcommand{\arraystretch}{1.3}
    \definecolor{hl}{rgb}{0.75,0.75,0.75}
\caption{Comparison of MFEA-RL with state-of-the-art EMT algorithms on the CEC17-MTSO for 30 independent runs}
\begin{center}
\resizebox{\textwidth}{!}{
\begin{tabular}{ccccccccc}
\hline
\textbf{Category} & \textbf{MFEA-RL} & \textbf{AT-MFEA} & \textbf{MFEA} & \textbf{MFEA-II} & \textbf{MTEA-SaO} & \textbf{MTES} & \textbf{BLKT-DE} & \textbf{MFEA-AKT} \\
\hline
P1-T1 &  \cellcolor{hl}\textbf{1.1587e-02 (1.24e-02)} & 2.7935e-01 (6.41e-02) - & 2.8606e-01 (1.03e-01) - & 6.5222e-01 (5.77e-02) - & 5.9571e-01 (9.26e-02) - & 6.2190e-01 (1.56e-01) - & 9.8086e-01 (5.41e-02) - & 2.5013e-01 (9.62e-02) - \\

P1-T2 &  \cellcolor{hl}\textbf{3.8071e+01 (4.70e+01)} & 3.2501e+02 (3.40e+01) - & 2.5383e+02 (4.52e+01) - & 3.6949e+02 (1.49e+01) - & 4.0895e+02 (2.37e+01) - & 3.7460e+02 (1.60e+01) - & 4.1541e+02 (2.15e+01) - & 1.6374e+02 (6.04e+01) - \\
\hline
P2-T1 &  \cellcolor{hl}\textbf{7.4762e-02 (1.57e-01)} & 1.6298e+00 (2.76e-01) - & 2.0952e+00 (3.16e-01) - & 3.0435e+00 (2.21e-01) - & 3.0315e+00 (1.86e-01) - & 2.9583e+00 (4.60e-01) - & 5.0312e+00 (6.28e-01) - & 2.8073e+00 (5.36e-01) - \\

P2-T2 &  \cellcolor{hl}\textbf{4.2680e+01 (6.80e+01)} & 2.1806e+02 (4.89e+01) - & 2.6163e+02 (4.33e+01) - & 3.6058e+02 (2.66e+01) - & 4.0769e+02 (2.83e+01) - & 3.7305e+02 (1.41e+01) - & 4.5906e+02 (3.27e+01) - & 1.7363e+02 (5.56e+01) - \\
\hline
P3-T1 &  \cellcolor{hl}\textbf{1.3280e-01 (3.78e-01)} & 1.9682e+01 (4.79e+00) - & 2.0612e+01 (1.24e-01) - & 2.1225e+01 (3.98e-02) - & 2.1221e+01 (3.21e-02) - & 2.1225e+01 (3.19e-02) - & 2.1221e+01 (4.73e-02) - & 2.0626e+01 (1.11e-01) - \\

P3-T2 &  \cellcolor{hl}\textbf{1.0721e-01 (3.37e-01)} & 1.4545e+03 (5.00e+02) - & 2.5324e+03 (4.00e+02) - & 1.3725e+03 (3.44e+02) - & 2.7168e+03 (4.78e+02) - & 1.4725e+04 (8.16e+02) - & 1.0242e+04 (8.97e+02) - & 2.4785e+03 (4.39e+02) - \\
\hline
P4-T1 &  \cellcolor{hl}\textbf{8.7441e+01 (7.09e+01)} & 4.2062e+02 (1.81e+01) - & 4.0230e+02 (6.73e+01) - & 4.1694e+02 (1.35e+01) - & 4.3317e+02 (3.07e+01) - & 4.5348e+02 (4.08e+01) - & 4.5679e+02 (3.30e+01) - & 3.6229e+02 (6.60e+01) - \\

P4-T2 &  \cellcolor{hl}\textbf{5.1931e+00 (3.12e+00)} & 5.3969e+00 (1.19e+00) = & 1.0011e+01 (4.91e+00) - & 4.6322e+00 (8.46e-01) = & 2.0042e+01 (6.09e+00) - & 7.5367e+01 (2.32e+01) - & 1.0893e+02 (4.50e+01) - & 3.9403e+00 (1.67e+00) = \\
\hline
P5-T1 &  \cellcolor{hl}\textbf{1.1041e+00 (5.65e-01)} & 2.1474e+00 (1.51e-01) - & 2.0500e+00 (3.36e-01) - & 2.9013e+00 (2.92e-01) - & 3.5592e+00 (2.44e-01) - & 2.8224e+00 (4.91e-01) - & 5.7169e+00 (9.35e-01) - & 3.1112e+00 (4.18e-01) - \\

P5-T2 &  \cellcolor{hl}\textbf{7.3155e+01 (5.69e+01)} & 7.0413e+02 (8.77e+01) - & 4.4219e+02 (9.00e+01) - & 2.1651e+03 (1.84e+03) - & 5.3118e+03 (3.37e+03) - & 2.2152e+03 (1.07e+03) - & 3.4814e+04 (1.91e+04) - & 4.6742e+02 (1.09e+02) - \\
\hline
P6-T1 &  \cellcolor{hl}\textbf{5.2115e-02 (6.03e-02)} & 2.6712e+00 (2.20e-01) - & 1.7055e+01 (7.01e+00) - & 3.1568e+00 (5.92e-01) - & 3.7228e+00 (5.38e-01) - & 3.2966e+00 (6.85e-01) - & 1.2279e+01 (6.56e+00) - & 2.3696e+00 (5.69e-01) - \\

P6-T2 &  \cellcolor{hl}\textbf{5.2813e-01 (3.38e-01)} & 1.1042e+00 (7.97e-01) - & 1.6104e+01 (7.32e+00) - & 2.0470e+00 (1.17e+00) - & 5.0413e+00 (1.36e+00) - & 2.9822e+00 (4.30e-01) - & 6.8019e+00 (2.96e+00) - & 2.7513e+00 (1.19e+00) - \\
\hline
P7-T1 &  \cellcolor{hl}\textbf{7.5903e+01 (5.97e+01)} & 8.6139e+02 (2.44e+02) - & 9.5127e+02 (4.90e+02) - & 2.2144e+03 (4.59e+02) - & 4.4234e+03 (2.06e+03) - & 2.1729e+03 (1.93e+03) - & 2.0057e+04 (1.36e+04) - & 7.3069e+02 (2.85e+02) - \\

P7-T2 &  \cellcolor{hl}\textbf{5.1010e+01 (6.53e+01)} & 3.6608e+02 (1.92e+01) - & 3.0786e+02 (5.75e+01) - & 3.8008e+02 (1.40e+01) - & 4.3550e+02 (2.92e+01) - & 3.7008e+02 (2.06e+01) - & 4.6359e+02 (3.34e+01) - & 2.2993e+02 (5.10e+01) - \\
\hline
P8-T1 &  \cellcolor{hl}\textbf{3.4946e-02 (4.33e-02)} & 3.5259e-01 (7.06e-02) - & 4.4557e-01 (8.08e-02) - & 3.4682e-01 (6.56e-02) - & 6.3879e-01 (7.73e-02) - & 7.5966e-01 (1.12e-01) - & 1.0351e+00 (1.09e-01) - & 2.6459e-01 (6.43e-02) - \\

P8-T2 &  \cellcolor{hl}\textbf{3.5827e+00 (1.73e+00)} & 7.7704e+00 (8.50e-01) - & 2.1072e+01 (5.78e+00) - & 1.2519e+01 (2.04e+00) - & 1.9865e+01 (4.08e+00) - & 9.8327e+00 (1.46e+00) - & 1.7165e+01 (3.39e+00) - & 1.4255e+01 (1.82e+00) - \\
\hline
P9-T1 &  \cellcolor{hl}\textbf{9.6509e+01 (6.90e+01)} & 4.1996e+02 (1.67e+01) - & 4.2118e+02 (5.55e+01) - & 4.2117e+02 (1.56e+01) - & 4.3469e+02 (2.67e+01) - & 4.2355e+02 (3.16e+01) - & 6.0979e+02 (1.15e+02) - & 4.1196e+02 (7.86e+01) - \\

P9-T2 &  \cellcolor{hl}\textbf{7.0700e+02 (1.17e+03)} & 1.3630e+03 (3.11e+02) - & 2.6591e+03 (3.30e+02) - & 1.3070e+03 (3.14e+02) - & 2.4032e+03 (4.84e+02) - & 1.7123e+04 (3.44e+02) - & 5.4072e+03 (1.16e+03) - & 2.5132e+03 (4.54e+02) - \\
\hline
+ / - / = & Base & 0 / 17 / 1 & 0 / 18 / 0 & 0 / 17 / 1 & 0 / 18 / 0 & 0 / 18 / 0 & 0 / 18 / 0 & 0 / 17 / 1 \\
\hline
\end{tabular}}
\label{tab1}
\end{center}
\end{table*}

\begin{table*}[h!]
\vspace{0.5em}
    \setlength{\tabcolsep}{1.5pt}
    \renewcommand{\arraystretch}{1.3}
    \definecolor{hl}{rgb}{0.75,0.75,0.75}
\caption{Comparison of MFEA-RL with state-of-the-art EMT algorithms on the WCCI20-MTSO for 30 independent runs}
\begin{center}
\resizebox{\textwidth}{!}{
\begin{tabular}{ccccccccc}
\hline
\textbf{Category} & \textbf{MFEA-RL} & \textbf{AT-MFEA} & \textbf{MFEA} & \textbf{MFEA-II} & \textbf{MTEA-SaO} & \textbf{MTES} & \textbf{BLKT-DE} & \textbf{MFEA-AKT} \\
\hline
20-P1-T1 & 6.1651e+02 (1.47e+00) &  \cellcolor{hl}\textbf{6.0149e+02 (1.05e+00)} + & 6.4396e+02 (8.20e+00) - & 6.0377e+02 (1.52e+00) + & 6.2355e+02 (5.64e+00) - & 6.0552e+02 (2.38e+00) + & 6.0618e+02 (3.33e+00) + & 6.1688e+02 (3.29e+00) = \\

20-P1-T2 & 6.1776e+02 (1.93e+00) &  \cellcolor{hl}\textbf{6.0148e+02 (9.70e-01)} + & 6.4534e+02 (6.76e+00) - & 6.0392e+02 (1.39e+00) + & 6.2457e+02 (6.00e+00) - & 6.0506e+02 (2.14e+00) + & 6.0719e+02 (3.07e+00) + & 6.1659e+02 (2.86e+00) = \\
\hline
20-P2-T1 &  \cellcolor{hl}\textbf{7.0001e+02 (5.13e-03)} & 7.0002e+02 (3.30e-02) = & 7.0000e+02 (1.51e-04) + & 7.0000e+02 (1.18e-04) + & 7.0000e+02 (6.11e-03) = & 7.0032e+02 (3.26e-01) - & 7.0000e+02 (1.49e-03) + & 7.0011e+02 (2.28e-01) = \\

20-P2-T2 &  \cellcolor{hl}\textbf{7.0000e+02 (1.42e-04)} & 7.0001e+02 (1.24e-02) = & 7.0002e+02 (2.95e-02) = & 7.0000e+02 (5.15e-03) + & 7.0001e+02 (6.08e-03) = & 7.0039e+02 (3.64e-01) - & 7.0000e+02 (1.32e-03) + & 7.0001e+02 (3.99e-03) = \\
\hline
20-P3-T1 &  \cellcolor{hl}\textbf{3.4695e+04 (1.62e+04)} & 2.3306e+06 (1.43e+06) - & 2.1792e+06 (1.18e+06) - & 1.9439e+06 (8.78e+05) - & 6.0571e+05 (2.16e+05) - & 1.1985e+07 (6.27e+06) - & 1.7335e+06 (7.01e+05) - & 4.1190e+05 (2.77e+05) - \\

20-P3-T2 &  \cellcolor{hl}\textbf{7.5921e+04 (3.58e+04)} & 2.4483e+06 (1.11e+06) - & 2.4015e+06 (1.05e+06) - & 1.8755e+06 (1.32e+06) - & 9.2503e+05 (3.90e+05) - & 1.4511e+07 (9.95e+06) - & 1.5688e+06 (1.16e+06) - & 4.1155e+05 (2.40e+05) - \\
\hline
20-P4-T1 & 1.3006e+03 (1.41e-01) & 1.3005e+03 (5.67e-02) = & 1.3006e+03 (1.20e-01) = & 1.3004e+03 (6.55e-02) = &  \cellcolor{hl}\textbf{1.3004e+03 (8.10e-02)} + & 1.3007e+03 (9.62e-02) = & 1.3006e+03 (8.60e-02) = & 1.3005e+03 (7.30e-02) = \\

20-P4-T2 & 1.3005e+03 (4.74e-02) & 1.3004e+03 (6.72e-02) = & 1.3004e+03 (7.06e-02) = & 1.3005e+03 (3.39e-02) = & 1.3004e+03 (6.21e-02) = &  \cellcolor{hl}\textbf{1.3004e+03 (5.32e-02)} = & 1.3006e+03 (9.35e-02) - & 1.3004e+03 (7.10e-02) = \\
\hline
20-P5-T1 & 1.5370e+03 (9.18e+00) & 1.5289e+03 (5.07e+00) + & 1.5359e+03 (9.87e+00) = & 1.5274e+03 (6.14e+00) = &  \cellcolor{hl}\textbf{1.5142e+03 (3.85e+00)} + & 1.5458e+03 (1.01e+01) = & 1.5307e+03 (5.59e+00) = & 1.5288e+03 (7.55e+00) + \\

20-P5-T2 & 1.5292e+03 (6.01e+00) & 1.5257e+03 (8.40e+00) = & 1.5342e+03 (9.64e+00) = & 1.5289e+03 (6.63e+00) = &  \cellcolor{hl}\textbf{1.5114e+03 (2.80e+00)} + & 1.5291e+03 (5.37e+00) = & 1.5317e+03 (6.35e+00) = & 1.5279e+03 (6.16e+00) = \\
\hline
20-P6-T1 &  \cellcolor{hl}\textbf{3.1645e+05 (1.73e+05)} & 1.3358e+06 (7.63e+05) - & 1.4911e+06 (1.29e+06) - & 7.2456e+05 (3.28e+05) - & 4.9806e+05 (2.68e+05) = & 2.3531e+07 (1.37e+07) - & 7.6785e+05 (3.31e+05) - & 8.4371e+05 (6.01e+05) - \\

20-P6-T2 & 1.1965e+06 (7.32e+05) & 1.1034e+06 (5.07e+05) = & 1.0622e+06 (7.71e+05) = & 6.3580e+05 (2.80e+05) = &  \cellcolor{hl}\textbf{2.4771e+05 (1.24e+05)} + & 1.6124e+07 (7.49e+06) - & 4.5854e+05 (2.77e+05) + & 6.5242e+05 (4.50e+05) + \\
\hline
20-P7-T1 & 2.9837e+03 (4.03e+02) &  \cellcolor{hl}\textbf{2.6927e+03 (2.60e+02)} = & 3.3282e+03 (2.83e+02) = & 2.6935e+03 (3.77e+02) = & 2.9587e+03 (3.93e+02) = & 3.5847e+03 (3.43e+02) - & 3.0083e+03 (3.80e+02) = & 2.9958e+03 (2.91e+02) = \\

20-P7-T2 &  \cellcolor{hl}\textbf{2.7729e+03 (2.27e+02)} & 3.0524e+03 (2.88e+02) - & 3.2201e+03 (2.79e+02) - & 2.9315e+03 (4.14e+02) = & 3.2059e+03 (3.83e+02) - & 3.5620e+03 (4.72e+02) - & 3.2150e+03 (2.78e+02) - & 3.2058e+03 (4.71e+02) - \\
\hline
20-P8-T1 &  \cellcolor{hl}\textbf{5.2000e+02 (1.68e-03)} & 5.2119e+02 (1.63e-02) - & 5.2031e+02 (9.05e-02) = & 5.2120e+02 (2.00e-02) - & 5.2118e+02 (4.97e-02) - & 5.2032e+02 (1.95e-01) = & 5.2120e+02 (4.79e-02) - & 5.2026e+02 (1.11e-01) = \\

20-P8-T2 & 5.2028e+02 (2.04e-01) & 5.2118e+02 (2.97e-02) - & 5.2030e+02 (9.79e-02) = & 5.2118e+02 (2.87e-02) - & 5.2117e+02 (5.98e-02) - &  \cellcolor{hl}\textbf{5.2000e+02 (8.88e-04)} + & 5.2120e+02 (3.22e-02) - & 5.2029e+02 (8.60e-02) = \\
\hline
20-P9-T1 & 8.3436e+03 (1.11e+03) & 1.4045e+04 (1.57e+03) - & 8.1960e+03 (8.62e+02) = & 1.3638e+04 (1.98e+03) - & 8.2357e+03 (2.32e+03) = &  \cellcolor{hl}\textbf{6.6014e+03 (1.11e+03)} + & 9.0832e+03 (2.24e+03) = & 7.9908e+03 (1.16e+03) = \\

20-P9-T2 &  \cellcolor{hl}\textbf{1.6208e+03 (8.42e-01)} & 1.6222e+03 (1.83e-01) - & 1.6212e+03 (4.12e-01) = & 1.6222e+03 (2.63e-01) - & 1.6218e+03 (8.65e-01) - & 1.6215e+03 (7.39e-01) = & 1.6225e+03 (3.65e-01) - & 1.6209e+03 (6.89e-01) = \\
\hline
20-P10-T1 &  \cellcolor{hl}\textbf{1.4391e+04 (8.21e+03)} & 2.3782e+04 (1.01e+04) = & 2.5073e+04 (1.13e+04) = & 2.6529e+04 (6.79e+03) = & 2.0617e+04 (6.80e+03) = & 1.2250e+06 (9.22e+05) - & 3.6535e+04 (2.88e+04) = & 3.3241e+04 (1.47e+04) - \\

20-P10-T2 & 1.8344e+06 (1.98e+06) & 1.3959e+06 (1.09e+06) = & 1.7609e+06 (1.62e+06) = & 3.1104e+06 (1.39e+06) - &  \cellcolor{hl}\textbf{4.1742e+05 (2.52e+05)} + & 1.4162e+07 (1.15e+07) - & 1.5338e+06 (1.02e+06) = & 2.7202e+06 (1.16e+06) - \\
\hline
+ / - / = & Base & 3 / 8 / 9& 1 / 6 / 13 & 4 / 8 / 8 & 5 / 8 / 7 & 4 / 10 / 6 & 5 / 8 / 7 & 2 / 6 / 12 \\
\hline
\end{tabular}}
\label{tab:comparison}
\end{center}
\end{table*}

We evaluate our algorithm on two widely used EMT benchmarks: CEC2017-MTSO \cite{cec_2017} and WCCI2020-MTSO. The evaluation budget is set to 50,000 and 200,000 for these benchmarks, respectively. CEC2017-MTSO consists of nine dual-task problems (P1–P9), categorized by task similarity into CI/PI/NI (complete/partial/no intersection) and HS/MS/LS (high/medium/low similarity). WCCI2020-MTSO extends the challenge with ten task pairs (20-P1 to 20-P10) derived from CEC2014, offering more complex combinations. These benchmarks provide diverse and representative scenarios for assessing algorithm robustness and adaptability. All experiments are implemented on the MTO platform proposed by Li et al. \cite{Mto_plat}.

\subsection{Comparison with Several State-of-the-art EMT Algorithms}

As shown in Tables \ref{tab1} and \ref{tab:comparison}, the experimental results compare the proposed MFEA-RL algorithm with several state-of-the-art EMT algorithms, including AT-MFEA \cite{at_mfea}, MFEA \cite{mfea}, MFEA-II \cite{mfea_ii}, MTEA-SaO \cite{sao}, MTES \cite{mtes}, BLKT-DE \cite{blktde}, and MFEA-AKT \cite{mfea_akt}.

\begin{figure*}[h!]
\vspace{0.8em}
\centerline{\includegraphics[width=7.2 in]{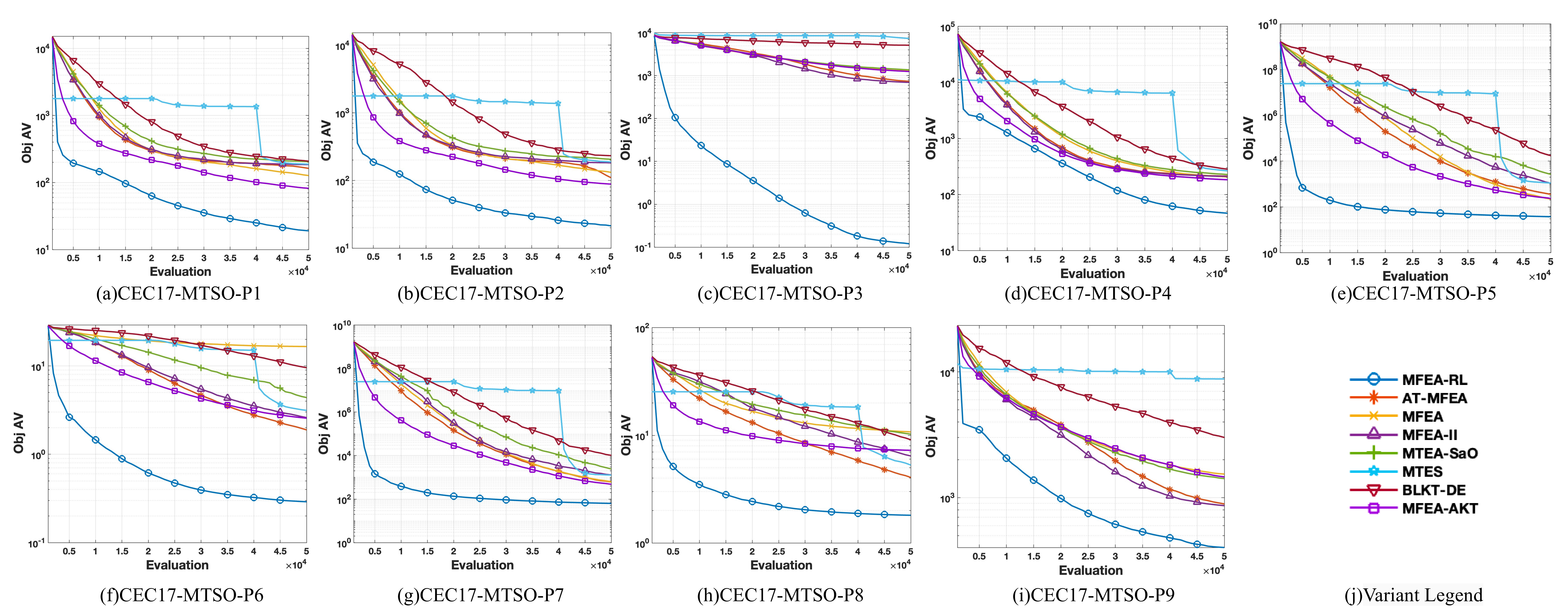}}
\caption{The convergence trends of the average objective value of MFEA-RL, AT-MFEA, MFEA, MFEA-II, MTEA-SaO, MTES, BLKT-DE and MFEA-AKT on CEC17-MTSO}
\label{convergence}
\end{figure*}

\begin{figure*}[h!]
\centerline{\includegraphics[width=7.1in]{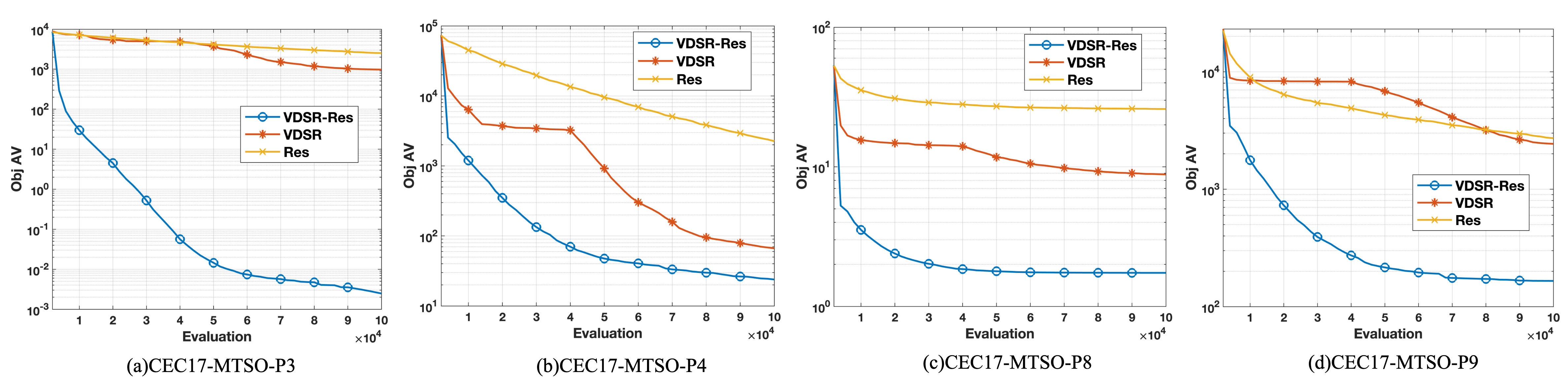}}
\caption{Convergence curves of the average objective value of VDSR-Res, VDSR and Res for representative tasks}
\label{fig_ablation}
\end{figure*}

On the CEC17-MTSO benchmark, MFEA-RL consistently outperforms the competing algorithms across nearly all task pairs, especially in the categories of CI-HS, CI-MS, PI-HS, PI-MS, PI-LS, NI-HS, and NI-MS, as evidenced by the lowest objective values (highlighted in bold). To further assess the statistical significance of these performance differences, Wilcoxon rank-sum tests were conducted at the 0.05 significance level. The statistical outcomes are presented in the last row of the table using the symbols "+/-/=" to indicate whether MFEA-RL performs better, worse, or comparably to other algorithms, respectively. Additionally, the values in parentheses indicate the standard deviation over 30 independent runs, reflecting performance stability. For the WCCI20-MTSO benchmark, MFEA-RL also demonstrates competitive performance across all task pairs. The tasks in this benchmark, derived from the CEC2014 single-objective optimization test set, present greater complexity and diverse task relationships, including both highly correlated and weakly correlated tasks. MFEA-RL achieves the best results on many tasks, as indicated by bolded values, further validating its robustness and adaptability. Statistical comparisons, reflected in the "+/-/=" row, highlight its superiority in handling high-complexity multitasking scenarios. 

MFEA-RL outperforms other algorithms on both the CEC17-MTSO and WCCI20-MTSO benchmarks due to its unique integration of residual learning and dynamic task adaptation, which effectively addresses the limitations of traditional multitasking optimization methods. Unlike algorithms such as MFEA and MFEA-II, which rely on fixed crossover and mutation strategies, MFEA-RL incorporates a VDSR model to transform individuals into high-dimensional residual representations, enabling the algorithm to capture complex inter-variable relationships. This capability allows MFEA-RL to better explore the decision space and transfer meaningful knowledge across tasks. Moreover, in contrast to AT-MFEA and MFEA-AKT, which use predefined or static skill factor assignment strategies, MFEA-RL employs a ResNet to dynamically assign skill factors, ensuring a more adaptive distribution of computational resources based on evolving task relationships. This feature is particularly critical in scenarios where task similarities vary, such as in WCCI20-MTSO's high-complexity tasks. Additionally, the use of random mapping for crossover further distinguishes MFEA-RL from approaches like BLKT-DE and MTES, which do not explicitly mitigate negative transfer. By mapping high-dimensional residual information back to the original search space, MFEA-RL avoids unintended interference between tasks and maintains solution diversity. The combined effect of these innovations not only enhances MFEA-RL's ability to suppress negative transfer but also ensures robust performance in tasks with weak correlations, as reflected in both benchmarks. These strengths make MFEA-RL a versatile and effective algorithm for handling diverse and challenging multitasking scenarios.

\subsection{Ablation Study on the Improved Crossover Operator and Skill Factor Assignment Strategy}

\begin{table*}[h!]
\vspace{0.5em}
\setlength{\tabcolsep}{2pt}
    \renewcommand{\arraystretch}{1.2}
\definecolor{hl}{rgb}{0.75,0.75,0.75}
\caption{Comparison of MFEA-RL with advanced algorithms on SCP CATEGORIES}
\begin{center}
\resizebox{\textwidth}{!}{ 
\begin{tabular}{cccccc}
\hline
\textbf{Category} & \textbf{MFEA-RL} & \textbf{DEORA} & \textbf{MTSRA} & \textbf{SHADE} & \textbf{MFMP} \\
\hline
SCP(29-32) & \cellcolor{hl}\textbf{5.8616e+01 (2.46e+00)}  & 2.7219e+02 (1.07e+01) - & 1.5869e+02 (6.90e+00) - & 1.4946e+02 (6.29e+00) - & 1.0436e+02 (3.25e+00) - \\
\hline
SCP(27-33) & \cellcolor{hl}\textbf{6.2231e+01 (1.23e+00)}  & 2.7472e+02 (7.95e+00) - & 1.6369e+02 (3.87e+00) - & 1.5060e+02 (3.49e+00) - & 1.0738e+02 (2.09e+00) - \\
\hline
SCP(29-32) & \cellcolor{hl}\textbf{6.8431e+01 (1.60e+00)}  & 2.7718e+02 (5.35e+00) - & 1.6553e+02 (3.68e+00) - & 1.5524e+02 (3.96e+00) - & 1.1042e+02 (1.27e+00) - 
\\
\hline
+ / - / = & Base & 0 / 3 / 0 & 0 / 3 / 0 & 0 / 3 / 0 & 0 / 3 / 0 \\
\hline
\end{tabular}
}
\label{tab_scp}
\end{center}
\end{table*}

\begin{figure*}[h!]
\vspace{0.3em}
\centerline{\includegraphics[width=5.5 in]{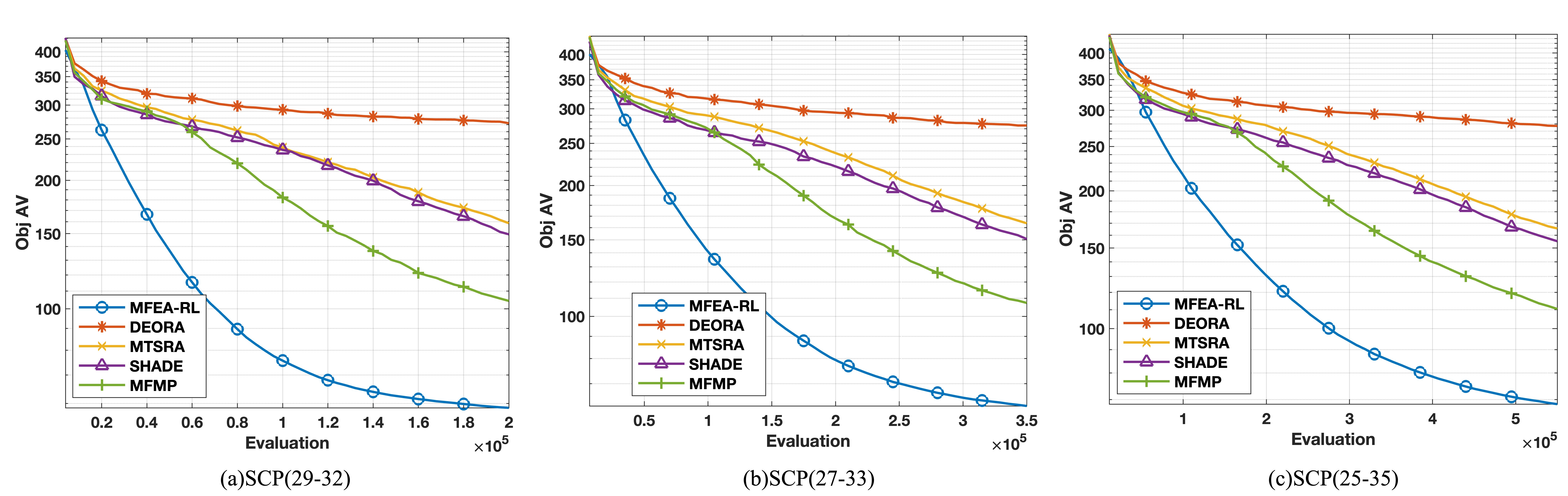}}
\caption{Convergence curves of algorithms on SCP problems}
\label{fig_scp}
\end{figure*}

Figure \ref{fig_ablation} illustrates the convergence curves for representative tasks (CEC17-MTSO-P3, P4, P8, and P9) in the ablation study. These results validate the synergistic effects of the improved crossover operator (VDSR + Random Mapping) and the dynamic skill factor assignment strategy (ResNet), labeled as VDSR-Res. The VDSR curve represents the performance when only the improved crossover operator is applied. The Res curve corresponds to the use of only the dynamic skill factor assignment strategy. The VDSR-Res curve combines both strategies. The experimental results demonstrate that VDSR-Res consistently outperforms the individual use of VDSR or ResNet across all tasks. 

In task P3, the VDSR-Res convergence curve shows a significant advantage from the initial evaluation phase, with a rapid decline compared to other strategies. This indicates its ability to effectively capture complex interactions between decision variables early in the optimization process and quickly approach the optimal solution. A similar trend is observed in task P4. Although ResNet alone exhibits some improvement, its combination with VDSR leads to complementary advantages, significantly enhancing both convergence speed and final performance. For tasks with higher complexity and weaker task correlations, such as P8 and P9, VDSR-Res not only achieves faster convergence but also obtains better final optimization results. This demonstrates that the improved crossover operator enhances the modeling capability for relationships between individuals through high-dimensional residual representations. The random mapping mechanism further reduces the risk of negative transfer, improving optimization efficiency and performance. Meanwhile, the dynamic skill factor assignment strategy learns the latent dependencies between tasks and flexibly adjusts resource allocation to meet the optimization demands of different tasks. This combined strategy is particularly effective in scenarios with weak task correlations, significantly enhancing the algorithm's robustness and applicability.

\subsection{Real World Problems and Applications}

To evaluate the potential of the proposed algorithm in practical applications, we utilized three multitasking scenarios for the Sensor Coverage Problem (SCP) \cite{scp} designed by Li et al. \cite{yc_scp} to assess the performance of MFEA-RL. Specifically, the multitasking SCP scenarios consist of 4, 7, and 11 tasks, denoted as SCP (29-32), SC (27-33), and SCP (25-35), respectively. The SCP aims to optimize sensor deployment to achieve optimal monitoring coverage of target areas, which is highly relevant to real-world applications. These scenarios not only involve high-dimensional optimization problems but also feature complex task dependencies and collaborative relationships, highlighting the challenges and complexities of multitasking optimization.

We included Li et al.'s algorithm, MTSRA, along with several multitasking optimization comparison algorithms they employed in this context, including SHADE \cite{shade}, MFMP \cite{mfmp}, and DEORA \cite{deora}, and compared their performance with the proposed MFEA-RL. As shown in Table \ref{tab_scp} and Figure \ref{fig_scp}, MFEA-RL demonstrated outstanding performance across all three SCP scenarios, particularly excelling in task information sharing and knowledge transfer efficiency, significantly outperforming the comparison algorithms. These results validate the robustness and adaptability of MFEA-RL in addressing complex multitasking optimization problems and provide strong support for its potential in practical applications.

\section{CONCLUSION}
This paper introduced the MFEA-RL, which leverages high- dimensional residual representations through a VDSR-inspired crossover operator and a dynamic skill factor assignment strategy via ResNet. These enhancements significantly improved the algorithm’s ability to model complex inter-variable relationships, mitigate negative transfer, and adapt to diverse task dependencies. The proposed method consistently outperformed state-of-the-art approaches across benchmarks, demonstrating faster convergence and superior solution quality. Ablation studies highlighted the complementary effects of the crossover operator and skill factor assignment, confirming their synergistic impact on performance and stability. Practical experiments validate the algorithm's effectiveness in tackling real-world challenges with complex inter-variable dependencies, highlighting its application potential.

In future research, we aim to enhance the interpretability of the proposed algorithm by incorporating explainable AI techniques. Analyzing the role of residual representations in decision-making, and providing insights into the dynamic skill factor allocation process. Such advancements will not only improve understanding of the algorithm's internal mechanisms but also facilitate its application to real-world problems requiring transparency and trust.

\end{document}